\documentclass[runningheads]{llncs}
\usepackage{graphicx}
\usepackage{tikz}
\usepackage{comment}
\usepackage{amsmath,amssymb} % define this before the line numbering.
\usepackage{color}

\usepackage[accsupp]{axessibility}  % Improves PDF readability for those with disabilities.

\usepackage{subfigure}
\usepackage{multirow}
\usepackage{booktabs}
\usepackage{relsize}
\usepackage{url}
\definecolor{darkgreen}{RGB}{60,200,60}

\newcommand{\mypar}[1]{\vspace{0.1cm}\noindent\textbf{#1}.}

\begin{document}
% \linenumbers
\pagestyle{headings}
\mainmatter
\def\ECCVSubNumber{100}  % Insert your submission number here

\title{ModSelect: Automatic Modality Selection for Synthetic-to-Real Domain Generalization} % Replace with your title

\begin{comment}
\titlerunning{ECCV-22 submission ID 4} 
\authorrunning{ECCV-22 submission ID 4} 
\author{Anonymous ECCV submission}
\institute{Paper ID 4}
\end{comment}
%******************

%\begin{comment}
\titlerunning{ModSelect: Automatic Modality Selection}
\author{Zdravko Marinov\and
Alina Roitberg \and \\
David Schneider\and Rainer Stiefelhagen}
\authorrunning{Zdravko Marinov et al.}
\institute{Institute for Anthropomatics and Robotics, Karlsruhe Institute of Technology
\email{firstname.lastname@kit.edu}}
%\end{comment}
%******************
\maketitle

\begin{abstract}
   Modality selection is an important step when designing multimodal systems, especially in the case of cross-domain activity recognition as certain modalities are more robust to domain shift than others. However, selecting only the modalities which have a positive contribution requires a systematic approach. %We tackle this problem by proposing a method for quantifying the contribution of individual modalities in late fusion models for human activity recognition. 
%
   %The contribution of a modality is defined as its influence on the performance change when it is included in the late fusion,
 %  
  % which can be estimated by validating the multimodal models using ground-truth labels. 
  %We mitigate this requirement
  We tackle this problem by proposing an unsupervised modality selection method (ModSelect), %which does not use test dataset labels in the process. 
  which does not require any ground-truth labels.
  We determine the correlation between the predictions of multiple unimodal classifiers and the domain discrepancy between their embeddings. Then, we systematically compute modality selection thresholds, which select only modalities with a high correlation and low domain discrepancy. We show in our experiments that %our unsupervised modality selection
  our method ModSelect chooses only modalities with positive contributions and consistently improves the performance on a \textsc{Synthetic$\rightarrow$Real} domain adaptation benchmark, narrowing the domain gap.
   \vspace*{-0.5cm}
   %We conduct extensive experiments on the \textsc{Synthetic$\rightarrow$Synthetic} and  \textsc{Synthetic$\rightarrow$Real} domain adaptation benchmarks and show how our unsupervised modality selection chooses modalities which improve the performance and bridge the domain gap.
   
   %%%%%%%%%%%%%%%%%%%%%%%%%%%%%%%%%%%
   % As an ablation study, we identify all the tendencies from our supervised and unsupervised methods by qualitatively inspecting 2D projections of the modality embeddings. 
   %
   %  depending on its performance gain on the test dataset.
   %
   % Archive
   %
   %Including a modality in the late fusion influences the performance on the test dataset. The modality contribution quantifies this performance change. 
   %
   % The contribution depends on whether including the modality in the late fusion leads to an improvement or a decline on the test dataset. 
   % We compute modality selection thresholds by utilizing the correlation between the predictions of multiple unimodal classifiers and the domain discrepancy between their embeddings, and select only modalities with a high correlation and low domain discrepancy.
\end{abstract}
\vspace*{-0.5em}

\section{Introduction}

Human activity analysis is vital for intuitive human-machine interaction, with applications ranging from driver assistance~\cite{martin2019drive} to smart homes and assistive robotics~\cite{roitberg2015multimodal}.
 Domain shifts, such as appearance changes, constitute a significant bottleneck for deploying such models in real-life.
For example, while simulations are an excellent way of economical data collection, a \textsc{Synthetic$\rightarrow$Real} domain shift leads to $\boldsymbol{>}\textbf{60}\boldsymbol{\%}$ drop in accuracy when recognizing daily living activities~\cite{roitberg2021let}.
\textit{Multimodality} is a way of mitigating this effect, since different types of data, such as RGB videos, optical flow and body poses, exhibit individual strengths and weaknesses.
%Nevertheless, a large portion of human activity research is dedicated to supervised methods operating on conventional RGB videos only, demonstrating remarkable results linked to the rise of deep learning in predefined constrained settings~\cite{kay2017kinetics,smaira2020short,xie2018rethinking,tran2018closer}. 
%However, the performance of such methods declines by a significant margin~\cite{roitberg2021let,choi2020shuffle,chen2019temporal} in case of a distributional shift, representing a significant bottleneck in real-life applications~\cite{sunderhauf2018limits,reiss2020deep}.
%\textit{Multimodality} is one way of mitigating this effect, since different types of data, such as RGB videos, optical flow and body poses, exhibit their individual strengths and weaknesses.
For example, models operating on body poses are less affected by appearance changes, as the relations between different joints are more stable given a good skeleton detector~\cite{das2020vpn,das2021vpn++}. 
RGB videos, in contrast, are more sensitive to domain shifts~\cite{sankaranarayanan2018learning,munro2020multi} but also convenient since they cover the complete scene and video is the most ubiquitous modality~\cite{chaquet2013survey,delaitre2010recognizing,sharma2012discriminative,sun2020human}.

Given the complementary nature of different data types (see Figure \ref{fig:modalities}), we believe, that  multimodality has a strong potential for improving domain generalization of activity recognition models, but \textit{which modalities to select} and \textit{how to fuse the information} become important questions.
%Combining different sources has strong potential for making activity recognition models more robust to distributional shifts, but \textit{which modalities to select} and  \textit{how to link the information} becomes an important question.
Despite its high relevance for applications, the question of modality selection has been often overlooked in this field.
The main goal of our work is to develop a systematic framework for studying the contribution of individual modalities in cross-domain human activity recognition.
We specifically focus on the \textsc{Synthetic$\rightarrow$Real} distributional shift~\cite{roitberg2021let}, which opens new doors for economical data acquisition but comes with an especially large domain gap.
We study five different modalities and examine how the prediction outcomes of  multiple unimodal classifiers correlate as well as the domain discrepancy between their embeddings.
We hope that our study will provide guidance for a better modality selection process in the future.

\textbf{Contributions and Summary.}
We aim to make a step towards effective use of multimodality in the context of cross-domain activity recognition, which has been studied mostly for RGB videos in the past~\cite{chaquet2013survey,delaitre2010recognizing,sharma2012discriminative,sun2020human}. 
This work develops the modality selection framework ModSelect for quantifying the importance of individual data streams and can be summarized in two major contributions. 
% (1) We propose a supervised workflow for selecting the most beneficial modalities for the specific task and examine the performance gain brought by five different modalities
(1) We propose a metric for quantifying the contribution of each modality for the specific task by calculating how the performance changes when the modality is included in the late fusion. Our new metric can be used by future research to justify decisions in modality selection. However, to estimate these performance changes, we use the ground-truth labels from the test data. (2) To detach ourselves from supervised labels, we propose to study the domain discrepancy between the embeddings and the correlation between the predictions of the unimodal classifiers of each modality. We use the discrepancy and the correlation to compute modality selection thresholds and show that these thresholds can be used to select only modalities with positive contributions w.r.t. our proposed metric in (1). Our unsupervised modality selection ModSelect can be applied in settings where no labels are present, e.g., in a multi-sensor setup deployed in unseen environments, where ModSelect would identify which sensors to trust.

%(2) To detach ourselves from supervised labels, we propose to study modality contributions in unsupervised fashion by looking at the domain discrepancy between the individual embedding.

%(3) Lastly, we conduct the first systematic study of  of different modality mixtures and fusion mechanisms, (\textit{e.g.}, early vs. late fusion) for cross-domain activity recognition.

%\lipcol{\lipsum[1-2]}

%\lipcol{\lipsum[1-2]}

\section{Related Work}
\subsection{Multimodal Action Recognition} 
The usage of multimodal data represents a common technique in the field of action recognition, and is applied for both: increasing performance in supervised learning as well as unsupervised representation learning. Multimodal methods for action recognition include approaches which make use of video and audio \cite{Korbar18,alwassel2019,Piergiovanni2020,Patrick2020,alayrac2020self}, optical flow \cite{Han20b,Piergiovanni2020}, text \cite{li2020} or pose information \cite{das2020vpn,das2021vpn++,rai2021cocon}.
Such methods can be divided into lower level \emph{early / feature} fusion  which is based on merging latent space information from multiple modality streams \cite{imran2020evaluating,zou2020deep,memmesheimer2020gimme,ahmad2019human,ahmad2020cnn,pham2021combining,gao2020listen,xiao2020audiovisual} and \emph{late / score fusion} which combines the predictions of individual classifiers or representation encoders either with learned fusion modules \cite{wang2019generative,wang2016exploring,ahmad2019human,zou2019wifi,kazakos2019epic,panda_adamml_2021} or with rule-based algorithms.

For this work, we focus on the latter, since the variety of early fusion techniques and learned late fusion impedes a systematic comparison, while rule-based late fusion builds on few basic but successful techniques such as averaging single-modal scores \cite{ye2015temporal,imran2016human,baradel2017human,ardianto2018multi,dawar2018convolutional,carreira2017quo,dhiman2020view,cai2021jolo,duan2021revisiting}, the max rule \cite{kamel2018deep,ardianto2018multi,dhiman2020view,rani2021kinematic}, product rule \cite{imran2016human,wang2017scene,kamel2018deep,Pushpajit2018,dawar2018data,zhao20193d,wei2019fusion,dhiman2020view,rani2021kinematic} or median rule. Ranking based solutions \cite{emerson2013original,Erp2002,ramanathan2019combining,cormack2009reciprocal} like Borda count are less commonly used for action recognition but recognised in other fields of computer vision.

\subsection{Modality Contribution Quantification}

While the performance contribution of modalities has been analyzed in multiple previous works, e.g., by measuring the signal-to-noise ratio between modalities \cite{wu2016novel}, determining class-wise modality contribution by learning an optimal linear modality combination \cite{kampman2018investigating,atrey2010multimodal} or extracting modality relations with threshold-based rules \cite{wang2021m2lens}, in-depth analysis of modality contributions in the field of action recognition remains sparse and mostly limited to small ablation studies. Metrics to measure data distribution distances like Maximuim Mean Discrepancy (MMD) or Mean Pairwise Distance (MPD) have been applied in fields like domain adaptation. MMD is commonly used to estimate and to reduce domain shift \cite{pan2010domain,long2013transfer,ghifary2014domain,wang2020rethink} and can be adapted to be robust against class bias, e.g. in the form of weighted MMD \cite{yan2017mind}, Mean Pairwise Distance (MPD) was applied to analyze semantic similarities of word embeddings, e.g., in \cite{elekes2017various}. 
In this work, we introduce a systematic approach for analyzing  modality contributions in the context of cross-domain activity recognition, which, to the best of out knowledge, has not been addressed in the past.

\subsection{Domain Generalization and Adaptation}
Both domain generalization and domain adaptation present strategies to learn knowledge from a source domain which is transferable to a given target domain. While domain adaptation allows access to data from the target domain to fulfill this task, either paired with labels~\cite{reiss2020deep,chen2020action,pan2020adversarial} or in the form of unsupervised domain adaptation \cite{busto2018open,chen2019temporal,reiss2020deep,chen2020action,pan2020adversarial,choi2020shuffle,choi2020unsupervised,song2021spatio}, domain generalization assumes an unknown target domain and builds upon methods which condition a neural network to make use of features which are found to be more generalizable \cite{yao2021videodg}, apply heavy augmentations to increase robustness \cite{yi2021benchmarking} or explore different methods of leveraging temporal data \cite{yi2021benchmarking}.
% \lipcol{\lipsum[1]}

\iffalse
\begin{itemize}
    \item It would be nice to explain Domain Generalization vs. UDA vs. DA and why DG is the most realistic/pragmatic scenario.
\end{itemize}
\fi
% 
%\subsection{Synthetic-to-Real Action Recognition}
%

%\lipcol{\lipsum[1-2]}

\section{Approach}

\begin{figure*}
    \centering
    \includegraphics[width=0.9\linewidth]{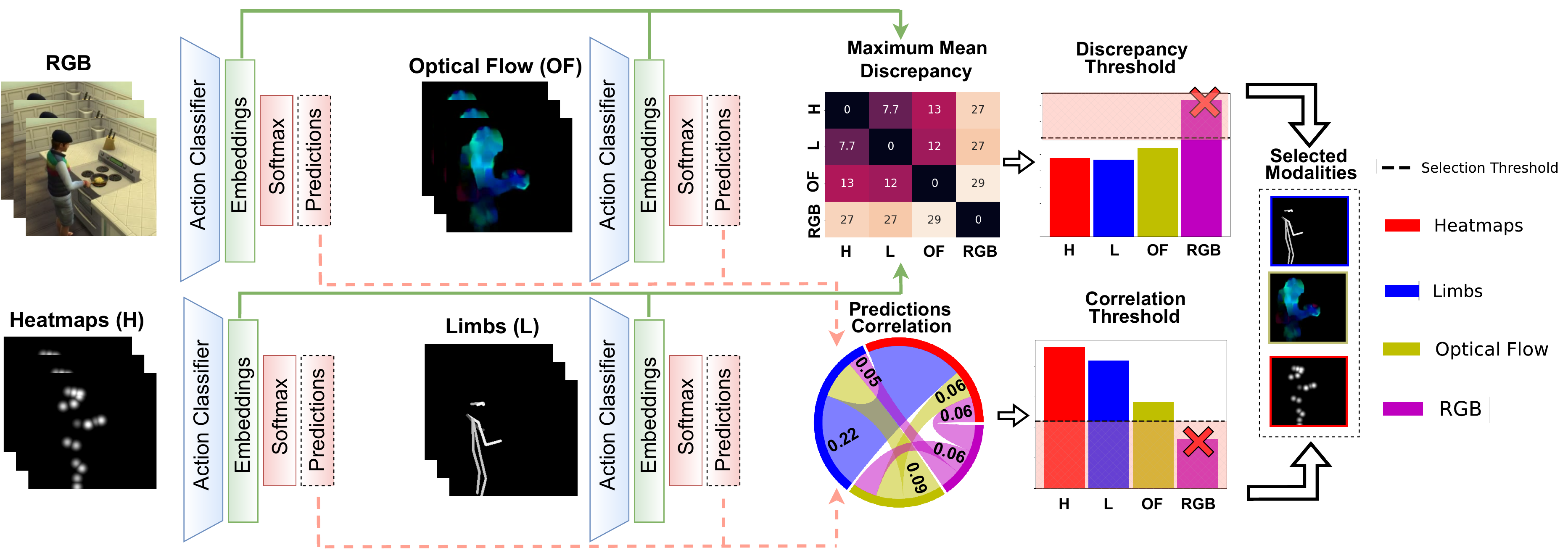}
    \caption{ModSelect: out approach for unsupervised modality selection which uses predictions correlations and domain discrepancy.}
    \label{fig:mod_select}
    \vspace*{-0.25cm}
\end{figure*}

Our approach consists of three main steps. (1) We extract multiple modalities and train a unimodal action recognition classifier on each modality. Afterwards, we evaluate all possible combinations of the modalities with different late fusion methods. We define the action recognition task in Section \ref{sec:ar_task}, the datasets we use in Section \ref{sec:datasets}, and the modality extraction and training in Section \ref{sec:modality_extraction_and_training}. (2) In Section \ref{sec:baseline_study}, we determine which modalities lead to a performance gain based on our evaluation results from (1). This establishes a baseline for the (3) third step (Section \ref{sec:unsupervised_quantification}), where we show how to systematically select these beneficial modalities in an unsupervised way with our framework ModSelect - without the need of labels nor evaluation results. We offer an optional notation table in the Supplementary for a better understanding of all of our equations.

We intentionally do not make use of learned late fusion techniques, such as \cite{wang2019generative,wang2016exploring,ahmad2019human,zou2019wifi,kazakos2019epic}, since such methods do not allow for comparing the contribution of individual modalities. Instead, a specific learned late fusion architecture could be better suited to some modalities in contrast to others, overshadowing a neutral evaluation. However, our work can be used to select modalities upon which such learned late fusion techniques can be designed. 
%Our approach consists of three main steps. The first step is to extract multiple modalities and train and evaluate multimodal action recognition classifiers on all possible modality combinations. We define the action recognition task in Section \ref{sec:ar_task}, the datasets we use in Section \ref{sec:datasets}, and the modality extraction and training in Section \ref{sec:modality_extraction_and_training}.
%
\subsection{Action Recognition Task} \label{sec:ar_task}
Our goal is to produce a systematic method for unsupervised modality selection in multimodal action recognition. More specifically, we focus on \textsc{Synthetic$\rightarrow$Real} domain generalization to show the need for a modality selection approach when a large domain gap is present. In this scenario an action classifier is trained only on samples from a \textsc{Synthetic} source domain $x_s \in X_s$ with action labels $y \in Y$. In domain generalization, the goal is to generalize to an unseen target domain $X_t$, without using any samples $x_t \in X_t$ from it during training. In our case, the target domain $X_t$ consists of \textsc{Real} data and the source and target data originate from distinct probability distributions $x_s \sim p_{synthetic}$ and $x_t \sim p_{real}$. The goal is to classify each instance $x_t$ from the \textsc{Real} target test domain $X_t$, which has a shared action label set $Y$ with the training set. To achieve this, we use the synthetic Sims4Action dataset~\cite{roitberg2021let} for training and the real Toyota Smarthome (Toyota)~\cite{das2019toyota} and ETRI-Activity3D-LivingLab (ETRI)~\cite{jang2020etri} as two separate target test sets. We also evaluate our models on the Sims4Action official test split~\cite{roitberg2021let} in our additional \textsc{Synthetic$\rightarrow$Synthetic} experiments.

\subsection{Datasets} \label{sec:datasets}
\iffalse
\begin{itemize}
    \item Go into detail about the datasets that we use - class distribution, scenes, subjects etc.
    \item Mention we use all three datasets for evaluation in Syn2Syn and Syn2Real to investigate different degrees of domain shift (ETRI $>$ Toyota $>$ Sims)
\end{itemize}
\fi
We focus on \textsc{Synthetic$\rightarrow$Real} domain generalization between the synthetic Sims4Action~\cite{roitberg2021let} as a training dataset and the real Toyota Smarthome~\cite{das2019toyota} and ETRI~\cite{jang2020etri} as test datasets. Sims4Action consists of ten hours of video material recorded from the computer game Sims 4, covering $10$ activities of daily living which have direct correspondences in the two \textsc{Real} datasets. Toyota Smarthome~\cite{das2019toyota} contains videos of 18 subjects performing 31 different everyday actions within a single apartment, and ETRI~\cite{jang2020etri} consists of 50 subjects performing 55 actions recorded from perspectives of home service robots in various residential spaces. However, we use only the 10 action correspondences to Sims4Action from the \textsc{Real} datasets for our evaluation.
%\lipcol{\lipsum[1]}
% 

\begin{figure}[b]
       \centering
       \includegraphics[width=0.7\linewidth]{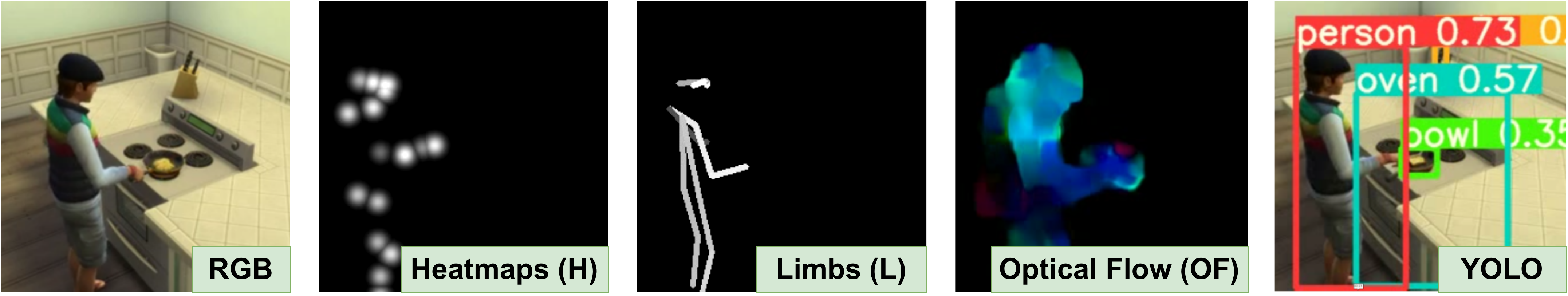}
       \caption{Examples of all extracted modalities. Note: the YOLO modality is represented as a vector \textbf{v}, which encodes distances to the person's detection (see Section \ref{sec:modality_extraction_and_training}).}
       \label{fig:modalities}
\end{figure}

\subsection{Modality Extraction and Training} \label{sec:modality_extraction_and_training}
We leverage the multimodal nature of actions to extract additional modalities for our training data, such as body pose, movement dynamics, and object detections. To this end, we utilize the RGB videos from the synthetic Sims4Action~\cite{roitberg2021let} to produce four new modalities - heatmaps, limbs, optical flow, and object detections. An overview of all modalities can be seen in Figure \ref{fig:modalities}.  

\textbf{Heatmaps and Limbs.} The heatmaps and limbs (H and L) are extracted via AlphaPose~\cite{fang2017rmpe,li2018crowdpose,xiu2018poseflow}, which infers 17 joint locations of the human body. The heatmaps modality $h(x,y)$ at pixel $(x,y)$ is obtained by stacking 2D Gaussian maps, which are centered at each joint location $(x_i, y_i)$ and each map is weighted by its detection confidence $c_i$ as shown in Equation \ref{eq:gaussian_heatmap}, where $\sigma=6$.
\begin{equation}
    h(x,y):=exp\bigg(\frac{-((x - x_i)^2 + (y - y_i)^2)}{2\sigma^2}\bigg) \cdot c_i
    \label{eq:gaussian_heatmap}
\end{equation}
The limbs modality is produced by connecting the joints with white lines and weighting each line by the smaller confidence of its endpoints. We weight both modalities by the detection confidences $c_i$ so that uncertain and occluded body parts are dimmer and have a smaller contribution.

\textbf{Optical Flow.} The optical flow modality (OF) is estimated via the Gunnar-Farneback method~\cite{farneback2003two}. The optical flow $of(x,y)$ at pixel $(x,y)$ encodes the \textit{magnitude} and \textit{angle} of the pixel intensity changes between two frames in the \textit{value} and \textit{hue} components of the HSV color space. The \textit{saturation} is used to adjust the visibility and we set it to its maximum value. The heatmaps, limbs, and optical flow are all \textit{image-based} and are used as an input to models which usually utilize RGB images. 

\textbf{Object Detections.} Our last modality (YOLO) consists of object detections obtained by YOLOv3~\cite{redmon2018yolov3}, which detects 80 different objects. Unlike the other modalities, we represent the detections as a vector, instead of an image. We show that such a simple representation achieves good domain generalization in our experiments. The YOLO modality for an image sample consists of a $k$-dimensional vector \textbf{v}, where \textbf{v$[i]$} corresponds to the reciprocal Euclidean distance between the person's and the $i^{\text{th}}$ object's bounding box centers, and $k=80$ is the number of detection classes. This way, objects closer to the person have a larger weight in \textbf{v} than ones which are further away. After computing the distances, \textbf{v} is normalized by its norm: $\textbf{v}\leftarrow \textbf{v}/||\textbf{v}||$. We denote the set of all modalities as $\mathcal{M}:=\{\text{H, L, OF, RGB, YOLO}\}$ and use the term $\mathcal{M}$ in our equations.

\textbf{Training.} We train unimodal classifiers on each modality and evaluate all possible modality combinations with different late fusion methods. We utilize 3D-CNN models with the S3D backbone~\cite{xie2018rethinking} for each one of the RGB, H, L, and OF modalities. The YOLO modality utilizes an MLP model as it is not image-based. We train all $5$ action recognition models end-to-end on Sims4Action~\cite{roitberg2021let}.

\textbf{Evaluation.} For our late fusion experiments, we combine the predictions of all unimodal classifiers at the class score level and obtain results for all $\sum_{i=1}^5{i \choose 5}=31$ modality combinations. We investigate $6$ late fusion strategies - Sum, Squared Sum, Product, Maximum, Median, and Borda Count \cite{ho1994decision,black1958theory}, which all operate on the class probability scores. Borda Count also uses the ranking of the class scores. For brevity, we refer to a late fusion of unimodal classifiers as a \textit{multimodal classifier} and present our late fusion results in Section~\ref{sec:ef_lf_results}.
\subsection{Quantification Study: Modality Contributions} \label{sec:baseline_study}
In this section, we propose how to quantify the contributions of each modality based on the performance of the models on the target test sets. To this end, we propose a with-without metric, which computes the average difference $f(m)$ of the performance of a multimodal classifier \textit{with} a modality $m$ to the performance \textit{without} it. Formally the contribution $f(m)$ of a modality $m$ is defined as:
\begin{equation}\label{eq:contribution}
    f(m) := \displaystyle \mathop{\mathbb{E}}_{C \in \mathcal{C}}[acc(C \cup \{m\}) - acc(C)]
\end{equation}
where $\mathcal{C}$ is the set of all modality combinations, $acc(C)$ is the test accuracy of the multimodal classifier with the modality combination $C$, and $m \in \mathcal{M}$. We compute $f(m)$ for all modalities based on the late fusion results listed in Section \ref{sec:ef_lf_results}. The contribution of each modality can be used to determine the modalities, which positively influence the performance on the test dataset $\mathcal{M}^+ \subseteq \mathcal{M}$.

%An advantage of using the with-without metric is that it is applicable to both early and late fusion models. To simulate the absence of a modality in the early fusion models we multiply its corresponding channels by $0$. Thus, the term $acc(C)$ in Equation \ref{eq:contribution} corresponds to the accuracy of the classifier trained on $C \cup \{m\}$ and evaluated on the test dataset with zero-valued channels for modality $m$. We compute $f(m)$ for all modalities based on our early- and late fusion results in Section \ref{sec:baseline_results}. We show that using the $f(m)$ metric leads to a separation between good and bad sources of information and to an overall higher performance.

%%%%%%%%%%%%%%%%%%%%%%%%%%%%%%%%%%%%%%%
\subsection{ModSelect: Unsupervised Modality Selection} \label{sec:unsupervised_quantification}
In this section, we introduce our method ModSelect for unsupervised modality selection. In this setting, we assume that we do not have any labels in the target test domain $X_t$. This is exactly the case for \textsc{Synthetic$\rightarrow$Real} domain generalization, where a model trained on simulated data is deployed in real-world conditions. In this case, the contribution of each modality cannot be estimated with Equation \ref{eq:contribution} as $acc(C)$ cannot be computed without ground-truth labels. Note that we do have labels in our test sets but we only use them in our quantification study and ignore them for our unsupervised experiments.

We propose ModSelect - a method for unsupervised modality selection based on the consensus of two metrics: (1) the correlation between the unimodal classifiers' predictions $\rho$ and (2) the Maximum Mean Discrepancy (MMD)~\cite{gretton2006kernel} between the classifiers' embeddings. We compute both metrics with our unimodal classifiers and propose how to systematically estimate modality selection thresholds. We show that the thresholds select the same modalities with positive contributions $\mathcal{M}^+$ as our quantification study in Section \ref{sec:baseline_study}. 

\textbf{Correlation Metric.} We define the predictions correlation vector $\boldsymbol{\rho}_{mn}$ between modalities $m$ and $n$ as:
\begin{equation}
    \boldsymbol{\rho}_{mn} := \frac{\mathbb{E}[(\textbf{z}_m - \boldsymbol{\mu}_{m}) \odot (\textbf{z}_n - \boldsymbol{\mu}_n) ]}{\boldsymbol{\sigma}_{m} \odot \boldsymbol{\sigma}_{n}}
\end{equation}
where $\textbf{z}_m, \textbf{z}_n$ are the softmax class scores of the action classifiers trained on modalities $m$ and $n$ respectively, $(\boldsymbol{\mu}_m, \boldsymbol{\sigma}_m), (\boldsymbol{\mu}_n, \boldsymbol{\sigma}_n)$ are the mean and standard deviation vectors of $\textbf{z}_m, \textbf{z}_n$, and $\odot$ is the element-wise multiplication operator. We define the predictions correlation $\rho(m,n)$ between two modalities $m$, $n$ as:
\begin{equation} \label{eq:pair_correlation}
    \rho(m,n):=\frac{1}{N}\sum_{i=1}^N \boldsymbol{\rho}_{mn}[i]
\end{equation}
where $N=10$ is the number of action classes. 

\textbf{MMD Metric.} We show that the distance between the distributions of the embeddings of two unimodal classifiers can also be used to compute a modality selection threshold. The MMD metric~\cite{gretton2006kernel} between two  distributions $P$ and $Q$ over a set $\mathcal{X}$ is formally defined as:
\begin{equation}
    MMD(P,Q) := ||\mathbb{E}_{X \sim P}[\varphi(X)] - \mathbb{E}_{Y \sim Q}[\varphi(Y)]||_{\mathcal{H}}
\end{equation}
where $\varphi: \mathcal{X}\mapsto \mathcal{H}$ is a feature map, and $\mathcal{H}$ is a reproducing kernel Hilbert space (RKHS)~\cite{gretton2012kernel,borgwardt2006integrating,gretton2006kernel}. For our empirical calculation of MMD between the embeddings $\textbf{h}_m, \textbf{h}_n$ of two modalities $m,n$ we set $\mathcal{X}=\mathcal{H}=\mathbb{R}^d$ and $\varphi(x)=x$:
\begin{equation} \label{eq:mmd_pairs}
    MMD(m,n):=||\mathbb{E}[\textbf{h}_m]- \mathbb{E}[\textbf{h}_n]||
\end{equation}
where $\textbf{h}_m, \textbf{h}_n$ are the embeddings from the second-to-last linear layer of the action classifiers for modalities $m$ and $n$ respectively, and $d$ is the embedding size. Note that using a linear feature map $\varphi(x)=x$ lets us determine only the discrepancy between the distributions' means. A linear mapping is sufficient to produce a good modality selection threshold, but one could also consider more complex alternatives, such as $\varphi(x)=(x,x^2)$ or a Gaussian kernel~\cite{gretton2012optimal}.

We make the following observations regarding both metrics for modality selection. Firstly, a high correlation between correct predictions is statistically more likely than a high correlation between wrong predictions, since there is only $1$ correct class and $N-1$ possibilities for error. We believe that a stronger correlation between the predictions results in a higher performance. Secondly, unimodal classifiers should have a high agreement on easy samples and a disagreement on difficult cases~\cite{ho1994decision}. A higher domain discrepancy between the classifiers' embeddings has been shown to indicate a lower agreement on their predictions~\cite{liang2021attention,zheng2015methodologies}, and hence, a decline in performance when fused. We therefore believe that good modalities are characterized by a low discrepancy and high correlation. % to other modalities' embeddings and predictions respectively.

\textbf{Modality Selection Thresholds.} After computing $\rho(m, n)$ and $MMD(m, n)$ for all pairs $(m,n)\in \mathcal{M}^2$, we systematically calculate modality selection thresholds for each metric $\rho$ and MMD. We consider two types of thresholds: (1) an aggregated threshold $\delta_{\textbf{agg}}$, which selects a set of \textit{individual modalities} $\mathcal{M}_{\textbf{agg}} \subseteq \mathcal{M}$, and (2) a pairs-threshold $\delta_{\textbf{pair}}$, which selects a set of \textit{modality pairs} $\mathcal{C}_{\textbf{pair}} \subseteq \mathcal{M}^2$.
% which separates modalities with a positive contribution $\mathcal{M}^+$ from ones with a negative $\mathcal{M}^-$, where $\mathcal{M}^+\cup \mathcal{M}^-=\mathcal{M}$.

% , where $\mathcal{M}^+\cup \mathcal{M}^-=\mathcal{M}=\{\text{H, L, OF, RGB, YOLO}\}$ and $\mathcal{M}^+ \cap \mathcal{M}^- = \emptyset$. 

\textbf{Aggregated Threshold $\delta_{\text{agg}}$.} For the first threshold, we aggregate the $\rho$ and $MMD$ values for a modality $m$ by averaging over all of its pairs: 
\begin{equation} \label{eq:agg_metrics}
\resizebox{0.6\textwidth}{!}{
    $\rho(m) := \mathlarger{\frac{1}{|\mathcal{M}|}}\displaystyle\sum_{n\in{\mathcal{M}}} \rho(m, n) \quad MMD(m) := \frac{1}{|\mathcal{M}|}\displaystyle\sum_{n\in{\mathcal{M}}} MMD(m, n)
    $
    }
\end{equation}
%The computed values from Equation \ref{eq:agg_metrics} for all modalities produce the sets $A_{\rho}$ and $A_{MMD}$.
Thus, we produce the sets $A_{\rho}:=\{\rho(m)|m \in \mathcal{M}\}$ and $A_{MMD}:=\{MMD(m) | m \in \mathcal{M}\}$. A simple approach would be to use the mean or median as a threshold for $A_{\rho}$ and $A_{MMD}$. However, such thresholds are sensitive to outliers (mean) or do not use all the information from the values (median). Additionally, one cannot tune the threshold with prior knowledge. To mitigate these issues, we propose to use the Winsorized Mean~\cite{huber1992robust,wilcox2003modern} $\mu_{\lambda}(A)$ for both sets, which is defined as:
\begin{equation} \label{eq:winsorized_mean}
    \mu_{\lambda}(A):= \lambda  a_{\lambda} + (1 - 2\lambda)\bar{a}_{\lambda} + \lambda a_{1-\lambda}
\end{equation}
where $a_{\lambda}$ is the $\lambda$-percentile of $A$, $\bar{a}_{\lambda}$ is the $\lambda$-trimmed mean of $A$, and $\lambda\in[0,0.5]$ is a ``trust'' hyperparameter. A higher $\lambda$ results in a lower contribution of edge values in $A$ and a bigger trust in values near the center. Therefore, we set $\lambda=0.2$ as we have $5$ modalities and expect to trust at least $3$. We compute two separate thresholds $\delta_{\textbf{agg}}^{\rho}:=\mu_{0.2}(A_{\rho})$ and $\delta_{\textbf{agg}}^{MMD}:=\mu_{0.2}(A_{MMD})$ and select the modalities $\mathcal{M}_{\textbf{agg}}$ as a consensus between the two metrics as:
\begin{equation} \label{eq:aggregated_consensus}
\resizebox{0.6\textwidth}{!}{
    $\mathcal{M}_{\textbf{agg}} := \{m \in \mathcal{M}| \rho(m) \geq \delta_{\textbf{agg}}^{\rho} \lor MMD(m) \leq \delta_{\textbf{agg}}^{MMD}\}  $
    }
\end{equation}
%\begin{equation}
%    \delta_{\textbf{agg}} := \frac{1}{2} (\mu_{\lambda}(A_\rho) + \mu_{\lambda}(A_{MMD}))
%\end{equation}
%

\textbf{Pairs-Threshold $\delta_{\textbf{pair}}$}. The second type of selection threshold skips the aggregation step of $\delta_{\textbf{agg}}$ and directly computes the Winsorized Means $\mu_{0.2}(\cdot)$ over the sets of all $\rho(m,n)$ and $MMD(m,n)$ values to obtain $\delta_{\textbf{pair}}^{\rho}$ and $\delta_{\textbf{pair}}^{MMD}$ respectively. This results in a selection of \textit{modality pairs}, rather than individual modalities as in Equation~\ref{eq:aggregated_consensus}. In other words, the $\delta_{\textbf{pair}}$ thresholds are suitable when one is searching for the \textit{best pairs} of modalities, and $\delta_{\textbf{agg}}$ for the \textit{best individual} modalities. The selected modality pairs $\mathcal{C}_{\textbf{pair}}$ with this method are:
\begin{equation} \label{eq:holistic_consensus}
    \resizebox{0.6\textwidth}{!}{
    $\mathcal{C}_{\textbf{pair}} := \{(m,n)\in \mathcal{M}^2 | \rho(m,n) \geq \delta_{\textbf{pair}}^{\rho} \lor MMD(m,n) \leq \delta_{\textbf{pair}}^{MMD}\}$
    }
\end{equation}

\textbf{Summary of our Approach.} A summary of our unsupervised modality selection method ModSelect is illustrated in Figure \ref{fig:mod_select}. We use the embeddings of unimodal action classifiers to compute the Maximum Mean Discrepancy (MMD) between all pairs of modalities. We also compute the correlation $\rho$ between the predictions of all pairs of classifiers. We systematically estimate thresholds for MMD and $\rho$ which discard certain modalities and select only modalities on which both metrics have a consensus. In the following Experiments \ref{sec:experiments} we show that the selected modalities $m\in \mathcal{M}$ with our method ModSelect are exactly the modalities with a positive contribution $f(m)>0$ according to Equation \ref{eq:contribution}, although our unsupervised selection does not utilize any ground-truth labels.

\section{Experiments}
\label{sec:experiments}
\subsection{Late Fusion: Results} \label{sec:ef_lf_results}

We evaluate our late fusion multimodal classifiers following the cross-subject protocol from~\cite{das2019toyota} for Toyota, the inter-dataset protocol from~\cite{kim2021action} for ETRI, and the official test split for Sims4Action from~\cite{roitberg2021let}. We follow the original \textsc{Sims4Action$\rightarrow$Toyota} evaluation protocol of~\cite{roitberg2021let} and utilize the mean-per-class accuracy (mPCA) as the number of samples per class are imbalanced in the \textsc{Real} test sets. The mPCA metric avoids bias towards overrepresented classes and is often used in unbalanced activity recognition datasets~\cite{das2019toyota,caba2015activitynet,martin2019drive,roitberg2021let}.

The results from our evaluation are displayed in Table~\ref{tab:lf_results}. The domain gap of transferring to  \textsc{Real} data is apparent in the drastically lower performance, especially on the ETRI dataset. Combinations including the H, L, or RGB modalities exhibit the best performance on the Sims4Action dataset, whereas OF and YOLO are weaker. However, combinations including the RGB modality seem to have an overall lower performance on the \textsc{Real} datasets, perhaps due to the large appearance change. Combinations with the YOLO modality show the best performance for both \textsc{Real} test sets. 
Inspecting the results in Table~\ref{tab:lf_results} is tedious and prone to misinterpretation or confirmation bias~\cite{nickerson1998confirmation}. It is also possible to overlook important tendencies. Hence, we show in Section~\ref{sec:baseline_results} how our quantification study tackles these problems by systematically disentangling the modalities with a positive contribution $\mathcal{M}^+$ from the rest $\mathcal{M}^-$.

\begin{table}
    \centering
    \caption{Results for the action classifiers trained on Sims4Action~\cite{roitberg2021let} in the mPCA metric. The late fusion results are averaged over the $6$  fusion strategies discussed in Section \ref{sec:modality_extraction_and_training}. H: Heatmaps, L: Limbs, OF: Optical Flow.}
    \scalebox{0.6}{
    \begin{tabular}{l|ccc||l|ccc}
    \toprule
        {} & \multicolumn{7}{c}{mPCA [\%]} \\ \cline{2-8}
        {} & \multicolumn{1}{c}{\textsc{Synthetic}}& \multicolumn{2}{c||}{\textsc{Real}} & {} & \multicolumn{1}{c}{\textsc{Synthetic}}& \multicolumn{2}{c}{\textsc{Real}} \\ \hline

        \multicolumn{1}{r|}{Test Set}& \multirow{1}{*}{Sims4Action~\cite{roitberg2021let}} & \multirow{1}{*}{Toyota~\cite{das2019toyota}} & \multirow{1}{*}{ETRI~\cite{jang2020etri}} &  \multicolumn{1}{r|}{Test Set} & \multirow{1}{*}{Sims4Action~\cite{roitberg2021let}} & \multirow{1}{*}{Toyota~\cite{das2019toyota}} & \multirow{1}{*}{ETRI~\cite{jang2020etri}} \\ \hline

        \multicolumn{1}{l|}{Modalities} & {} & {} & {} & \multicolumn{1}{l|}{Modalities}  & {} & {} & {} \\ \hline

       % {} & {} & {} & {} & {} \\ \hline
         
         H & \multicolumn{1}{c}{$71.38$} & \multicolumn{1}{c}{$20.23$} & \multicolumn{1}{c||}{$12.12$}  & H L OF RGB  & $96.95$  &  $26.00$  & $16.22$ \\
         L & \multicolumn{1}{c}{$75.09$} & \multicolumn{1}{c}{$22.00$} & \multicolumn{1}{c||}{$12.88$} & H L OF YOLO &   $90.05$ &   $28.98$ &   $20.27$ \\ 
         OF & \multicolumn{1}{c}{$44.50$} & \multicolumn{1}{c}{$21.34$} &
         \multicolumn{1}{c||}{$9.28$} & H L RGB YOLO &   $92.88$ &   $25.36$ &   $18.55$ \\ 
         RGB & \multicolumn{1}{c}{$61.79$} & \multicolumn{1}{c}{$13.74$} &
         \multicolumn{1}{c||}{$9.37$} & H OF RGB YOLO &   $91.14$ &  $26.10$ &   $17.20$ \\
         YOLO & \multicolumn{1}{c}{$50.54$} & \multicolumn{1}{c}{$26.08$} & \multicolumn{1}{c||}{$38.25$} & L OF RGB YOLO &   $92.56$ &  $26.12$ &   $18.56$ \\  \hline
         %%%%%%%%%%%%%%%%%%%%%%%%%%%%%%%
         H L &  $91.48$  & $23.44$  & $16.95$ & H L OF  & $94.03$  & $26.70$  & $17.26$ \\
         H OF  & $89.86$  & $25.97$  & $15.10$ & H L RGB  & $95.72$  & $23.25$  & $15.78$ \\
         H RGB  & $94.44$  & $21.38$  & $13.91$ & H OF RGB & $95.41$  & $23.73$  & $13.79$ \\
         L OF  & $90.84$  & $25.60$  & $15.90$ & L OF RGB  & $95.61$  & $24.09$  & $15.49$ \\
         L RGB  & $94.42$  & $20.58$  & $15.46$ & H L YOLO  & $86.57$ & $25.80$  & $20.46$ \\
         OF RGB  & $86.15$  & $18.50$ &  $12.85$ & H OF YOLO &   $82.97$ & $29.48$ &   $19.91$ \\ 
         %%%%%%%%%%%%%%%%%%%%%%%%%%%%%%%%%%
         H YOLO  & $74.86$  & $25.51$  & $20.55$ & H RGB YOLO &   $88.79$ & $23.27$ &   $17.77$ \\
         L YOLO  & $80.25$ & $24.77$  & $19.91$ & L OF YOLO &   $86.31$ &   $28.51$ &   $20.68$ \\
         OF YOLO & $62.21$  & $27.56$  & $20.34$ & L RGB YOLO &   $90.59$ &   $23.29$ &   $19.09$ \\
         RGB YOLO  & $76.23$  & $17.29$  & $19.41$ & OF RGB YOLO &   $82.37$ &   $21.15$ &   $16.92$ \\ \hline
         %%%%%%%%%%%%%%%%%%%%%%%%%%%%%%%%%

         %%%%%%%%%%%%%%%%%%%%%%%%%%%%%%%%%%%%
         H L OF RGB YOLO &   $94.27$ &   $27.52$ &   $18.53$  & {} & {} & {} & {} \\
         \bottomrule
    \end{tabular}
    }
    \label{tab:lf_results}
\end{table}

\subsection{Quantification Study: Results} \label{sec:baseline_results}

\begin{figure*}[b]
    \centering
\subfigure{\includegraphics[width=0.3\linewidth]{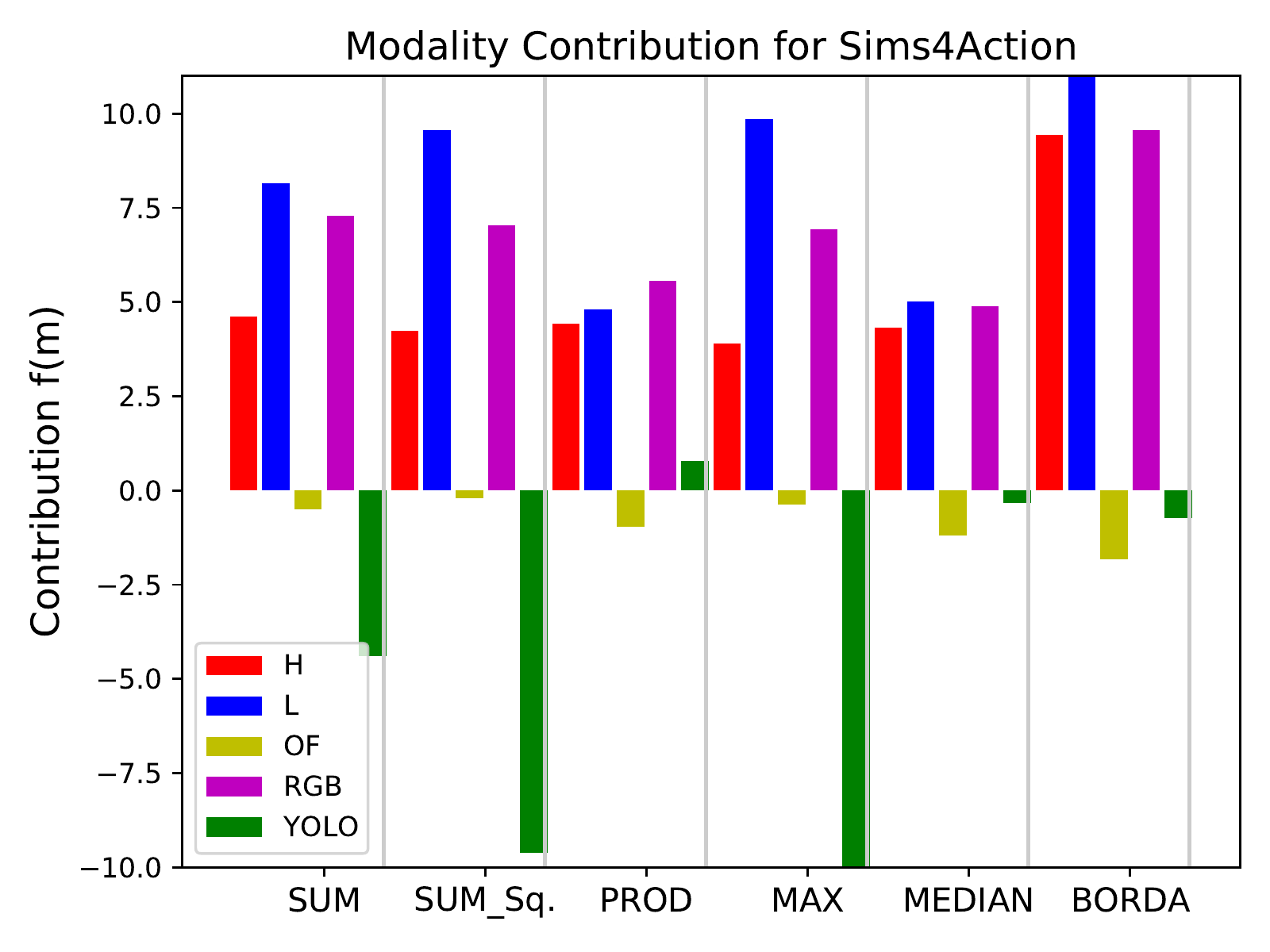}
  }
\subfigure{\includegraphics[width=0.3\linewidth]{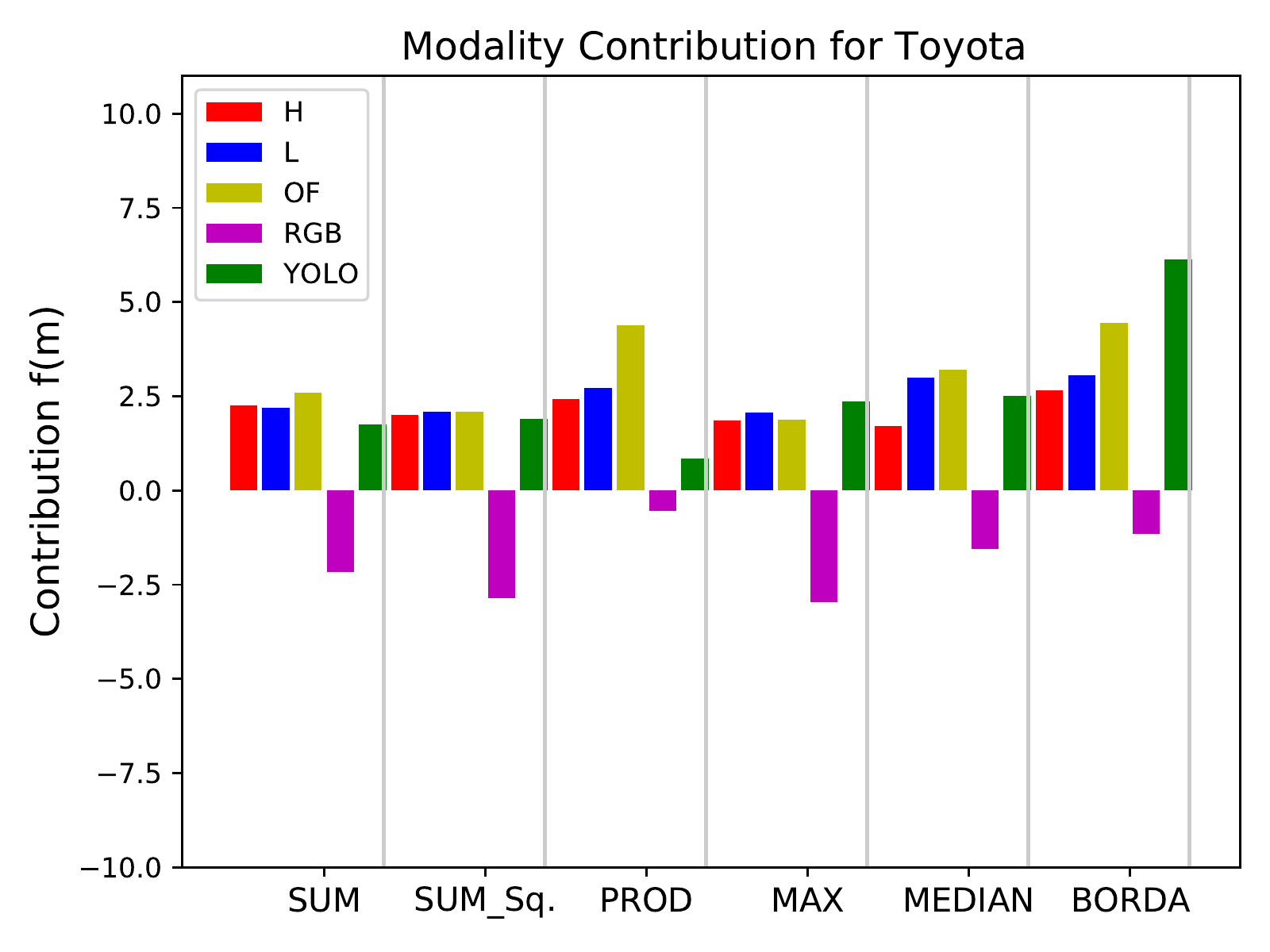}
  }
\subfigure{\includegraphics[width=0.3\linewidth]{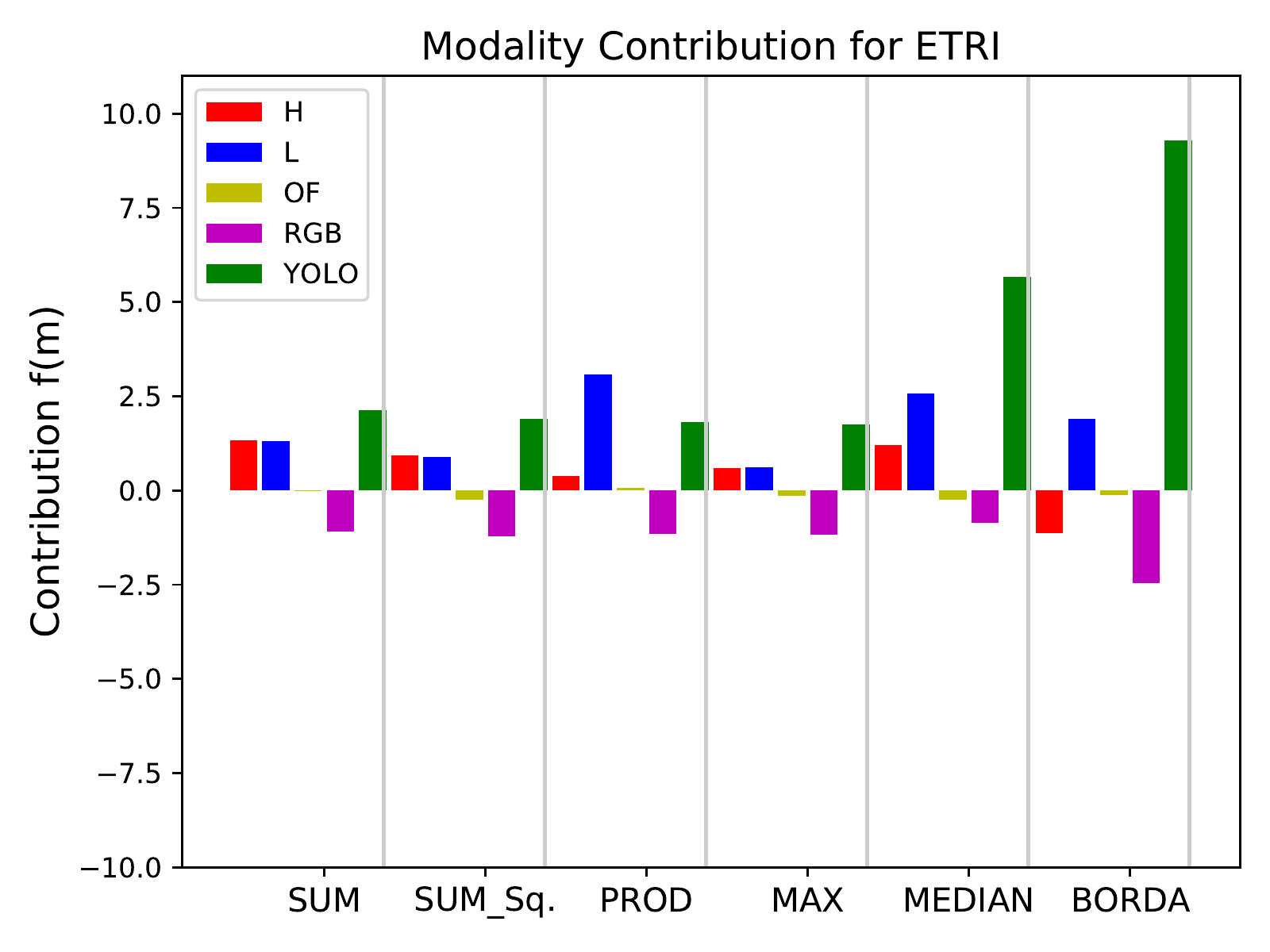}
  }
\iffalse
\begin{subfigure}{\linewidth}
  \includegraphics[width=0.3\linewidth]{figures/supervised_quantification/sims_contributions.pdf}
  \label{fig:4}
\end{subfigure}
\begin{subfigure}{\linewidth}
  \includegraphics[width=0.3\linewidth]{figures/supervised_quantification/toyota_contributions.pdf}
  \label{fig:baseline_5}
\end{subfigure}
\begin{subfigure}{\linewidth}
  \includegraphics[width=0.3\linewidth]{figures/supervised_quantification/etri_contributions.pdf}
  \label{fig:baseline_6}
\end{subfigure}
\fi

\caption{Quantification Study: Quantification of the contribution of each modality for $6$ late fusion methods and on three different test sets. The height of each bar corresponds to the contribution value $f(m)$ which is computed with Equation~\ref{eq:contribution}.}
\label{fig:baseline_results}
\end{figure*}

We use the results from Table \ref{tab:lf_results} for the $acc(\cdot)$ term in Equation \ref{eq:contribution} and compute the contribution of each modality $f(m)$. We do this for all late fusion strategies and all three test datasets and plot the results in Figure \ref{fig:baseline_results}. The \textsc{Synthetic$\rightarrow$Real} domain gap is clearly seen in the substantial difference in the height of the bars in the test split of Sims4Action~\cite{roitberg2021let} compared to the \textsc{Real} test datasets. The limbs and RGB modalities have the largest contribution on Sims4Action~\cite{roitberg2021let}, followed by the heatmaps. The only modalities with negative contributions are the optical flow and YOLO, where YOLO reaches a drastic drop of over $(-10\%)$ for the Squared Sum and Maximum late fusion methods. We conclude that YOLO and optical flow have a negative contribution on Sims4Action~\cite{roitberg2021let}.

The results on the \textsc{Real} test datasets Toyota~\cite{das2019toyota} and ETRI~\cite{jang2020etri} show different tendencies. The contributions are smaller due to the domain shift, especially on the ETRI dataset. The RGB modality has explicitly negative contributions on both \textsc{Real} datasets. We hypothesize that this is due to the appearance changes when transitioning from synthetic to real data. Apart from RGB, optical flow also has a consistently negative contribution on the ETRI dataset. An interesting observation is that the domain gap is much larger on ETRI than on Toyota. A reason for this might be that Roitberg et al.~\cite{roitberg2021let} design Sims4Action specifically as a \textsc{Synthetic$\rightarrow$Real} domain adaptation benchmark to Toyota Smarthome~\cite{das2019toyota}, e.g., in Sims4Action the rooms are furnished the same way as in Toyota Smarthome. Our results indicate that optical flow has a negative contribution in ETRI, whereas RGB is negative in both \textsc{Real} datasets. The average contribution of each modality over the 6 fusion methods is in Table \ref{tab:merged_tables}\textbf{(a)}.
\begin{table}[b]
    \centering
    \caption{\textbf{(a)} Average contribution $f(m)$ over the $6$ late fusion methods of each modality $m$. Negative contributions are colored in \textcolor{red}{red}. \textbf{(b)} Aggregated prediction correlation $\rho(m)$ values for each modality $m$ on the three test datasets and the aggregated thresholds $\delta_{\textbf{agg}}^{\rho}$. Values \textbf{below} the threshold are colored in \textcolor{red}{red}. \textbf{(c)} Aggregated $MMD(m)$ values for each modality $m$ on the three test datasets and the aggregated thresholds $\delta_{\textbf{agg}}^{MMD}$. Values \textbf{above} the threshold are colored in \textcolor{red}{red}.}
    \label{tab:merged_tables}
        \scalebox{0.75}{

    \begin{tabular}{l|ccccc||ccccc|c||cccc|c}
    \toprule
        Test Dataset & \multicolumn{5}{c||}{ \textbf{(a)} Contribution $f(m)$} & \multicolumn{5}{c|}{\textbf{(b)} Aggregated $\rho(m)$} & {} & \multicolumn{4}{c|}{\textbf{(c)} Aggregated $MMD(m)$} & {} \\ \hline
         {} & H & L & OF & RGB & YOLO & H & L & OF & RGB & YOLO & $\delta_{\textbf{agg}}^{\rho}$ & H & L & OF & RGB & $\delta_{\textbf{agg}}^{MMD}$  \\ \hline
        Sims4Action~\cite{roitberg2021let} & 4.37 & 8.86 & \color{red}{-0.74} & 6.98 & \color{red}{-2.57} & 0.57 & 0.55 & \color{red}{0.38} & 0.50 & \color{red}{0.37} & 0.40 & 9.49 & 8.12 & \color{red}{13.07} & 9.92 & 10.15 \\ 
        Toyota~\cite{das2019toyota} & 2.14 & 2.46 & 2.90 & \color{red}{-1.86} & 2.13 & 0.23 & 0.21 & 0.14 & \color{red}{0.08} & 0.14 & 0.10 & 11.93 & 11.47 & 13.34 & \color{red}{20.79} & 14.38 \\ 
        ETRI~\cite{jang2020etri} & 0.76 & 1.60 & \color{red}{-0.13} & \color{red}{-1.17} & 2.02 & 0.14 & 0.14 & \color{red}{0.06} & \color{red}{0.05} & 0.13 & 0.08 & 17.84 & 17.76 & \color{red}{22.04} & \color{red}{24.91} & 20.64 \\ 
        \bottomrule
    \end{tabular}}
\end{table}

\subsection{Results from ModSelect: Unsupervised Modality Selection}

Table \ref{tab:merged_tables}\textbf{(a)} shows which modalities have a negative contribution on each target test dataset. However, to estimate these values we needed the performance, and hence, the labels for the target test sets. In this section, we show how to select only the modalities with a positive contribution without using any labels.

\begin{figure*}[t]
    \centering % <-- added
\iffalse
\begin{subfigure}{\linewidth}
  \includegraphics[width=0.3\linewidth]{figures/correlation_chords/chord_Sims.pdf}
  \label{fig:chord_1}
\end{subfigure}%\hfil % <-- added
\begin{subfigure}{\linewidth}
  \includegraphics[width=0.3\linewidth]{figures/correlation_chords/chord_toyota.pdf}
  \label{fig:chord_2}
\end{subfigure}%\hfil % <-- added
\begin{subfigure}{\linewidth}
  \includegraphics[width=0.3\linewidth]{figures/correlation_chords/chord_ETRI.pdf}
  \label{fig:chord_3}
\end{subfigure}
\fi 
\subfigure{\includegraphics[width=0.3\linewidth]{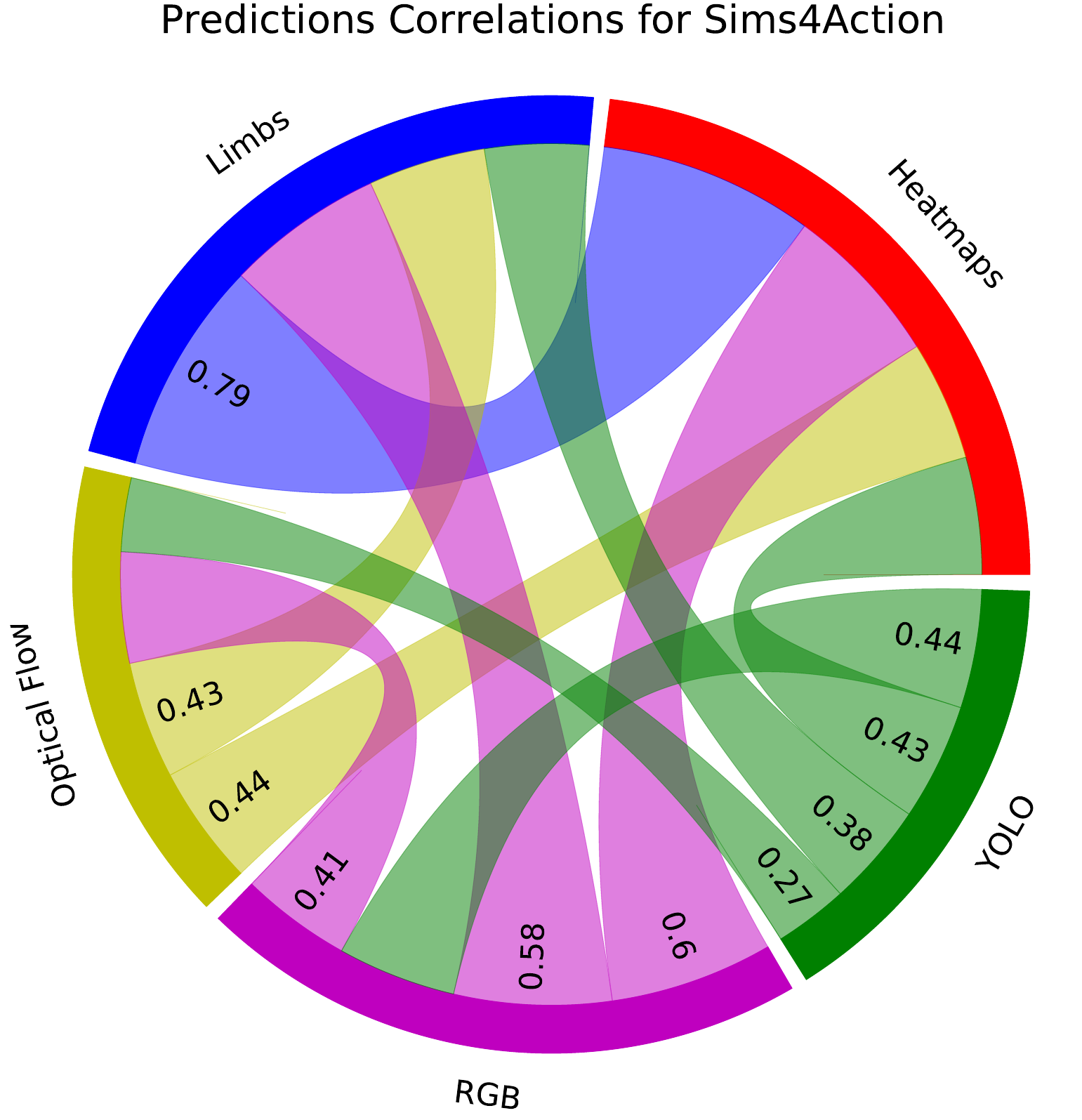}
  }
  \subfigure{\includegraphics[width=0.3\linewidth]{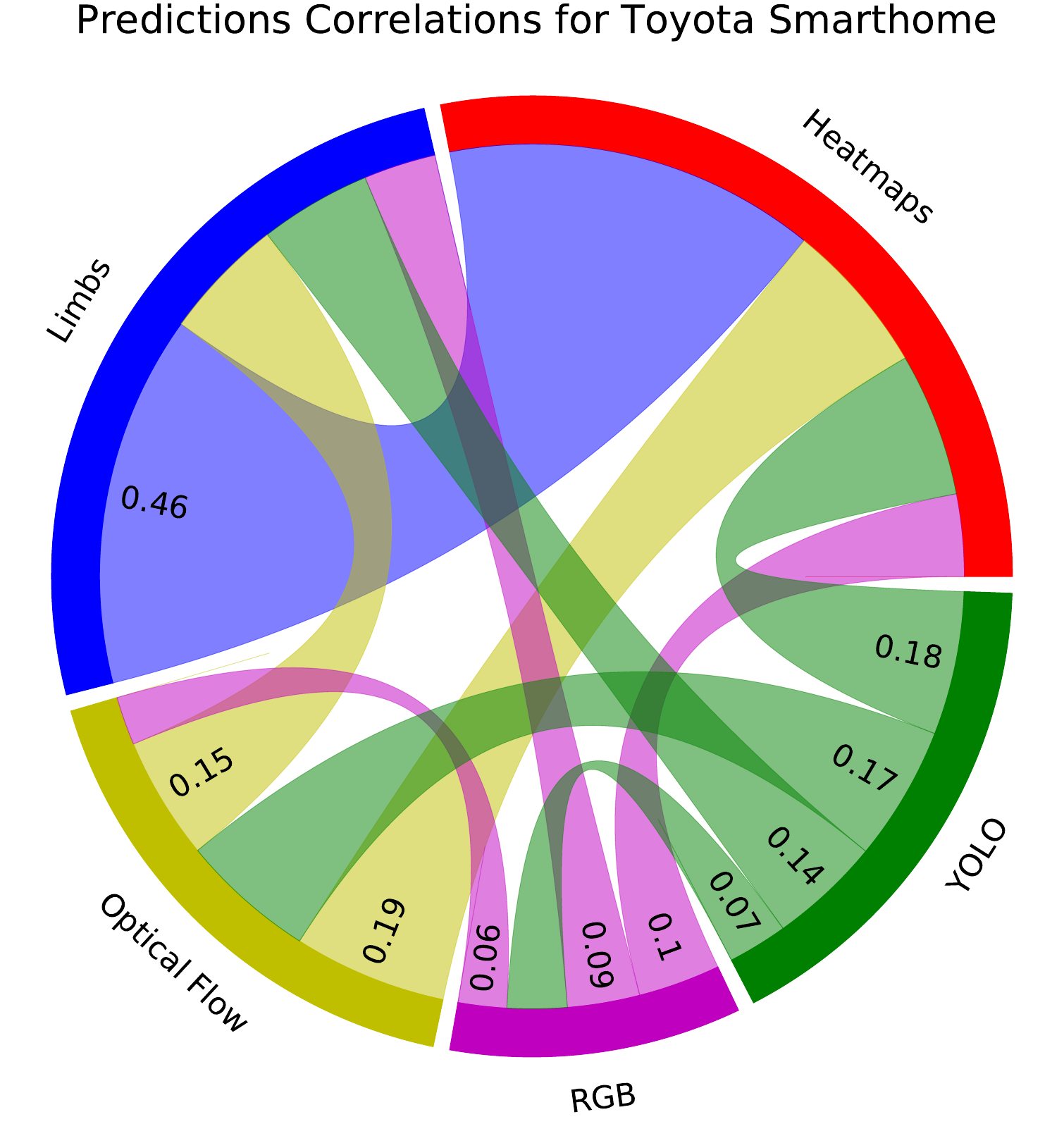}
  }
  \subfigure{\includegraphics[width=0.3\linewidth]{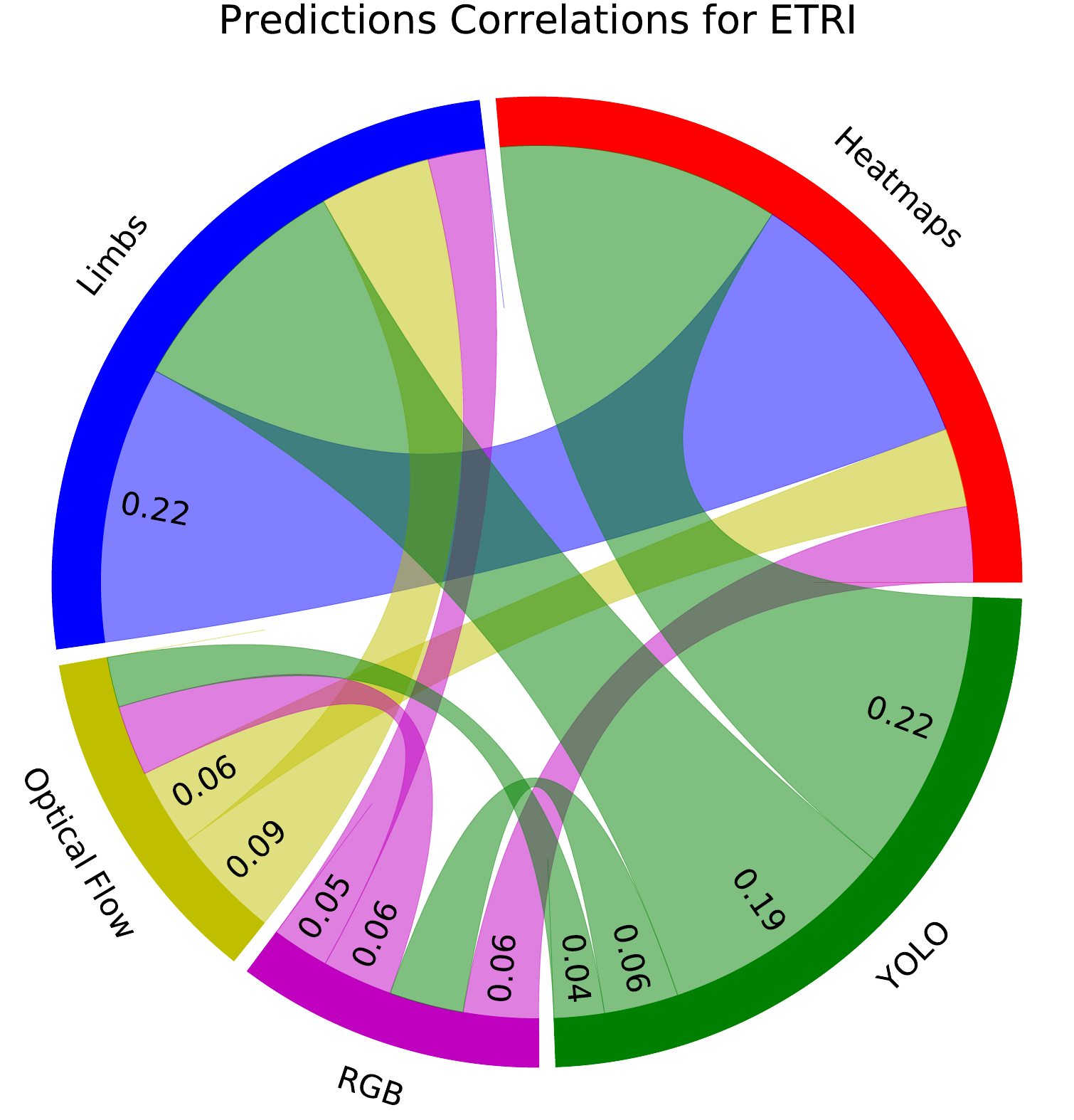}
  }

%\vspace*{-1.5em}
\caption{Chord plots of the prediction correlations $\rho(m,n)$ for all modality pairs $(m,n)$. The thickness of each arch corresponds to the correlation $\rho(m,n)$ between its two endpoint modalities $m$ and $n$. Each value is computed according to Equation~\ref{eq:pair_correlation}.}
\label{fig:chord_plots}
\vspace*{-0.25cm}

\end{figure*}

\textbf{Predictions Correlation.} We utilize the predictions correlation metric $\rho(m,n)$ and compute it for all modality pairs using Equation~\ref{eq:pair_correlation}. The results for all datasets are illustrated in the chord diagrams in Figure \ref{fig:chord_plots}. The chord diagrams allow us to identify the same tendencies, which we observed in our quantification study. Each arch connects two modalities $(m,n)$ and its thickness corresponds to the value $\rho(m,n)$. The YOLO modality has the weakest correlations on Sims4Action~\cite{roitberg2021let}, depicted in the thinner green arches. Optical flow also exhibits weaker correlations compared to the heatmaps, limbs, and RGB. We see significantly thinner arches for the RGB modality in both \textsc{Real} test datasets, and for optical flow in ETRI, which matches our results in Table~\ref{tab:merged_tables}\textbf{(a)}.

However, simply inspecting the chord plots is not a systematic method for modality selection. Hence, we first compute the aggregated correlations $\rho(m)$ with Equation~\ref{eq:agg_metrics}, which constitute the set $A_{\rho}$. Then, we compute the aggregated threshold $\delta_{\textbf{agg}}^{\rho}:=\mu_{0.2}(A_{\rho})$ using the $0.2$-Winsorized Mean~\cite{wilcox2003modern} from Equation~\ref{eq:winsorized_mean}. We compute these terms for each test dataset and obtain three thresholds. The aggregated correlations $\rho(m)$ and their thresholds $\delta_{\textbf{agg}}^{\rho}$ for each test dataset are illustrated in Table~\ref{tab:merged_tables}\textbf{(b)}. The modalities underneath the thresholds are exactly the ones with negative contributions from Table~\ref{tab:merged_tables}\textbf{(a)}.

\begin{figure}[t]
       \centering
       \includegraphics[width=0.95\linewidth]{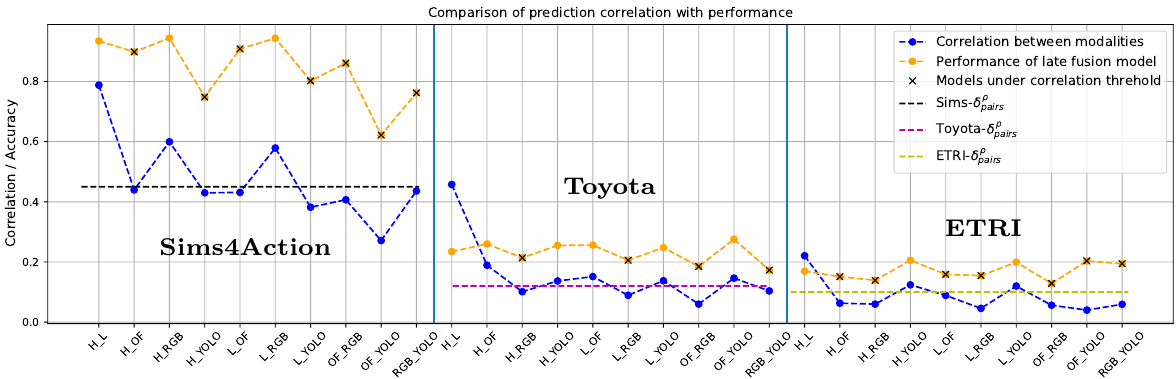}
       \vspace*{-0.5cm}
       \caption{Prediction correlations $\rho(m,n)$ between all modality pairs  and the late fusion accuracy of the bi-modal action classifiers. The pairs-thresholds $\delta_{\textbf{pair}}^{\rho}$ are depicted as dashed lines. Pairs under the thresholds are crossed out with an $\boldsymbol{\times}$ in the yellow line.}
       \label{fig:holistic_threshold_rho}
       \vspace*{-0.25cm}
\end{figure}

We also show that applying the pairs-threshold $\delta_{\textbf{pair}}^{\rho}$ leads to the same results. We skip the aggregation step of $\delta_{\textbf{agg}}^{\rho}$ and compose the $A_{\rho}$ set out of the $\rho(m,n)$ values, i.e. we focus on \textit{modality pairs} instead of \textit{individual modalities}. We compute the threshold $\delta_{\textbf{pair}}^{\rho}:=\mu_{0.2}(A_{\rho})$ again with the $0.2$-Winsorized Mean~\cite{wilcox2003modern} from Equation~\ref{eq:winsorized_mean}. The correlation values $\rho(m,n)$ as well as the accuracies of all bi-modal action classifiers from Table~\ref{tab:lf_results} are shown in Figure~\ref{fig:holistic_threshold_rho}. The pairs-thresholds $\delta_{\textbf{pair}}^{\rho}$ for each test set are drawn as dashed lines and divide the modality pairs into two groups. The modality pairs below the thresholds are marked with an $\boldsymbol{\times}$ so that it is possible to identify which pairs are selected.

\iffalse
\begin{figure}[b]
    \centering
    \includegraphics[width=0.8\linewidth]{figures/correlation_thresholds.pdf}
    \vspace*{-1em}
    \caption{Aggregated correlation $\rho(m)$ for each modality on all test datasets and aggregated thresholds $\delta_{\textbf{agg}}^{\rho}$ as dashed lines.}
    \label{fig:corr_thresholds}
    \vspace*{-1em}
\end{figure}
\fi
%\input{tables/correlations}

For Sims4Action~\cite{roitberg2021let}, optical flow (OF) and RGB show a negative $f(m)$ in Table~\ref{tab:merged_tables}\textbf{(a)} and are also discarded by $\delta_{\textbf{agg}}^{\rho}$. Figure~\ref{fig:holistic_threshold_rho} shows that all modality pairs containing either OF or RGB are below the threshold, i.e. $\delta_{\textbf{pair}}^{\rho}$ discards the same modalities as $\delta_{\textbf{agg}}^{\rho}$ for Sims4Action. The same is true for the \textsc{Real} test datasets, where all models containing RGB are discarded in Toyota and ETRI, as well as OF for ETRI. Another observation is that the majority of "peaks" in the yellow \textit{accuracy} lines coincide with the peaks in the blue \textit{correlation} lines. This result is in agreement with our theory that a high correlation between correct predictions is statistically more likely than a high correlation between wrong predictions, since there is one only correct class and multiple incorrect ones. Moreover, while we do achieve the same results with both thresholds, we recommend using $\delta_{\textbf{agg}}^{\rho}$ when discarding an entire input modality, e.g. a faulty sensor in a multi-sensor setup, and $\delta_{\textbf{pair}}^{\rho}$ when searching for the \textit{best synergies} from all modality combinations.

\textbf{Domain Discrepancy.} The second metric we use to discern the contributing modalities from the rest is the Maximum Mean Discrepancy (MMD)~\cite{gretton2006kernel} between the embeddings of the action classifiers. Note that the YOLO modality is not included in this experiment as its MLP model's embedding size is different that the other $4$ \textit{image-based} modalities. While MMD is widely used as a loss term for minimizing the domain gap between source and target domains~\cite{wang2020rethink,yan2017mind,chen2019graph}, a large MMD is also associated with a decline in performance in fusion methods~\cite{liang2021attention,zheng2015methodologies}. To utilize the MMD metrics to separate the modalities, we first compute $MMD(m,n)$ for all modality pairs and the aggregated discrepancies $MMD(m)$ with Equations~\ref{eq:mmd_pairs} and~\ref{eq:agg_metrics}. We then compute the pairs- $\delta_{\textbf{pair}}^{MMD}$ and aggregated $\delta_{\textbf{agg}}^{MMD}$ thresholds with the $0.2$-Winsorized Mean~\cite{wilcox2003modern} from Equation~\ref{eq:winsorized_mean} the same way as we did for the predictions'
correlations $\rho$. 

\begin{figure}[b]
    \centering % <-- added
\iffalse
\begin{subfigure}{\linewidth}
  \includegraphics[width=0.18\linewidth]{figures/mmd_matrices/mmd_Sims_Sims_cropped.pdf}
  \label{fig:mmd_1}
\end{subfigure} \hspace{15.0mm} % <-- added
\begin{subfigure}{\linewidth}
  \includegraphics[width=0.18\linewidth]{figures/mmd_matrices/mmd_Toyota_Toyota_cropped.pdf}
  \label{fig:mmd_2}
\end{subfigure} \hspace{15.0mm} % <-- added
\begin{subfigure}{\linewidth}
  \includegraphics[width=0.18\linewidth]{figures/mmd_matrices/mmd_ETRI_ETRI_cropped.pdf}
  \label{fig:mmd_3}
\end{subfigure}
\fi  

\subfigure{\includegraphics[width=0.18\linewidth]{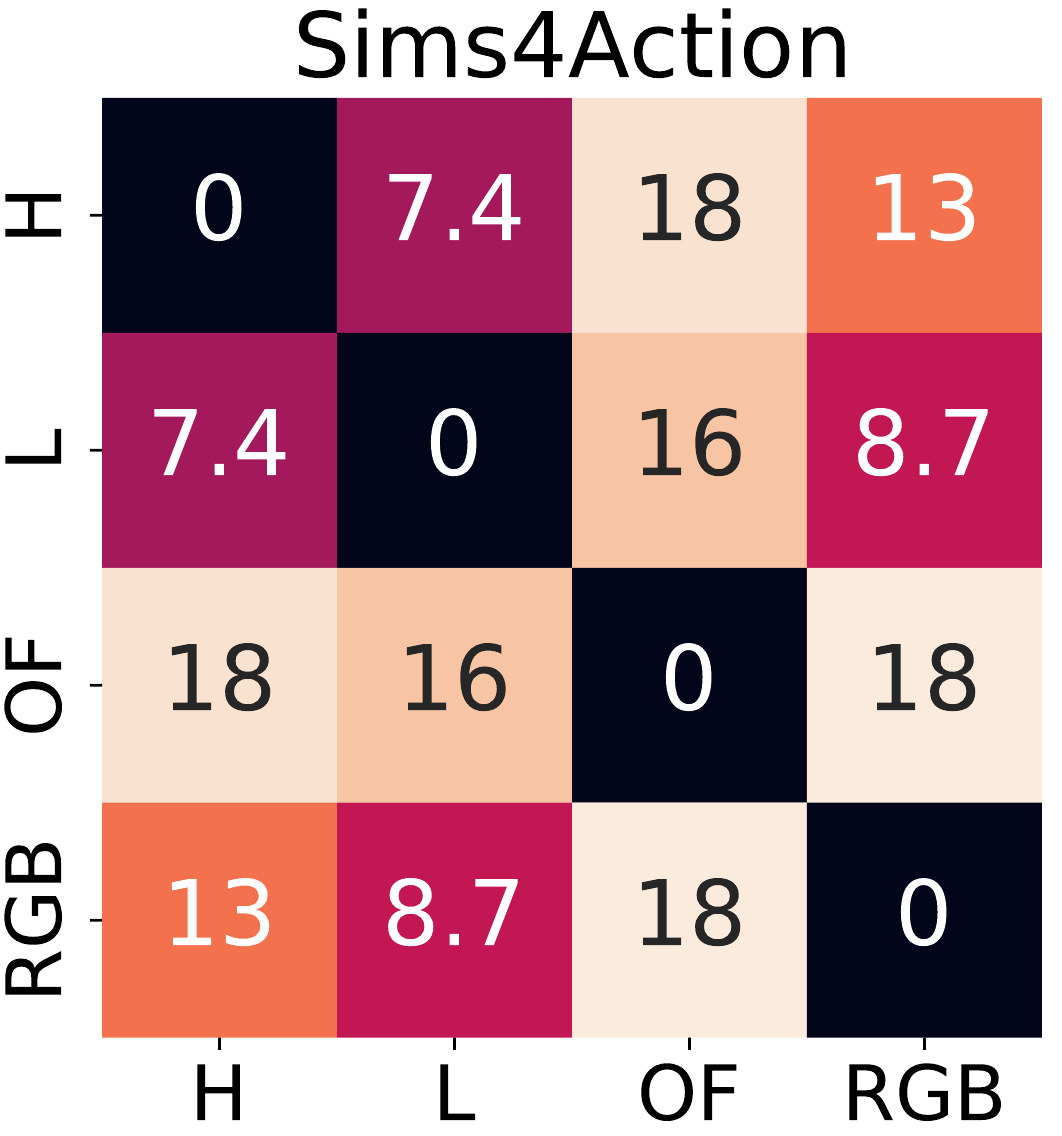} \hspace{15.0mm} % <-- added
  }
\subfigure{\includegraphics[width=0.18\linewidth]{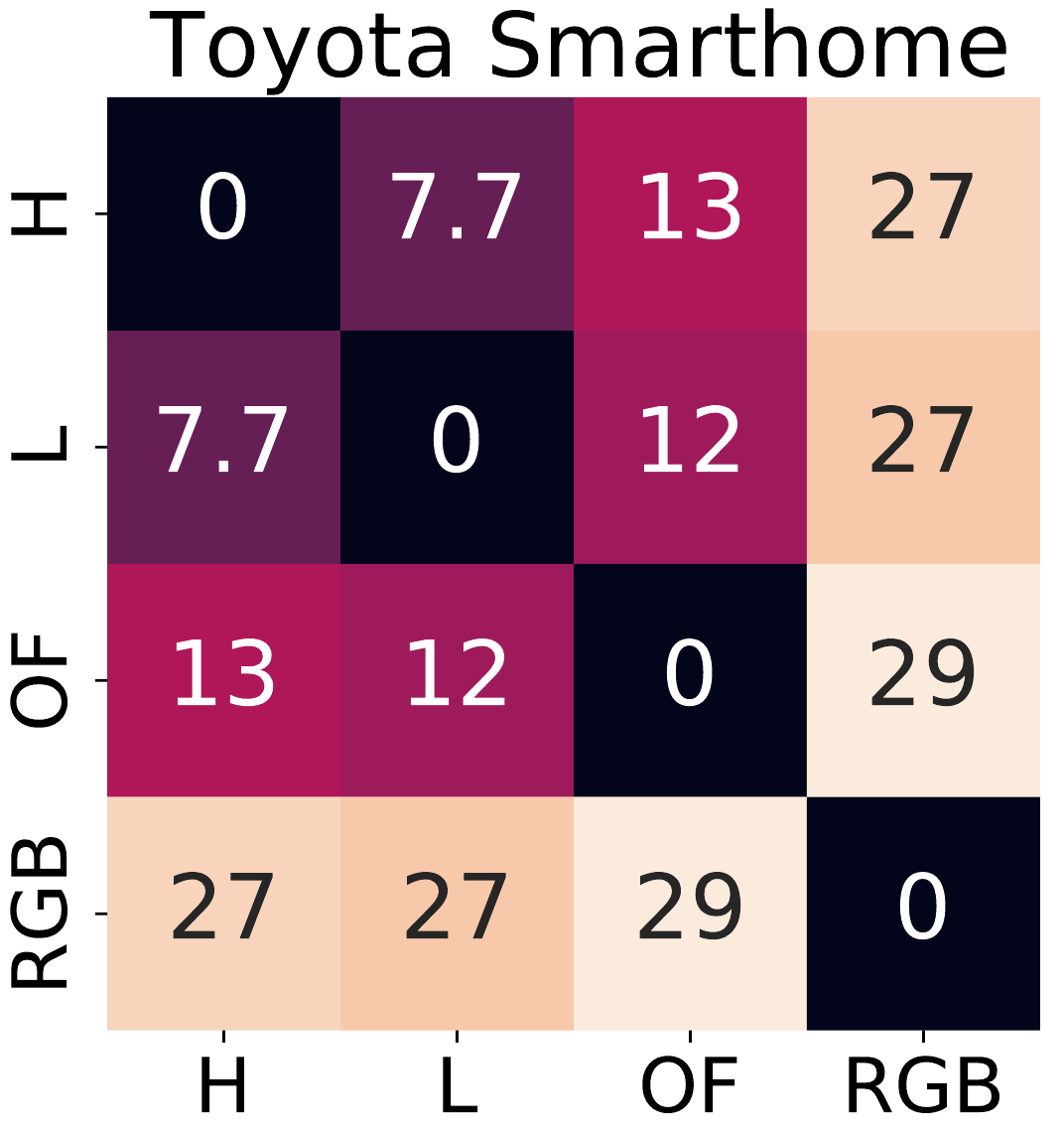} \hspace{15.0mm} % <-- added
  }
\subfigure{\includegraphics[width=0.18\linewidth]{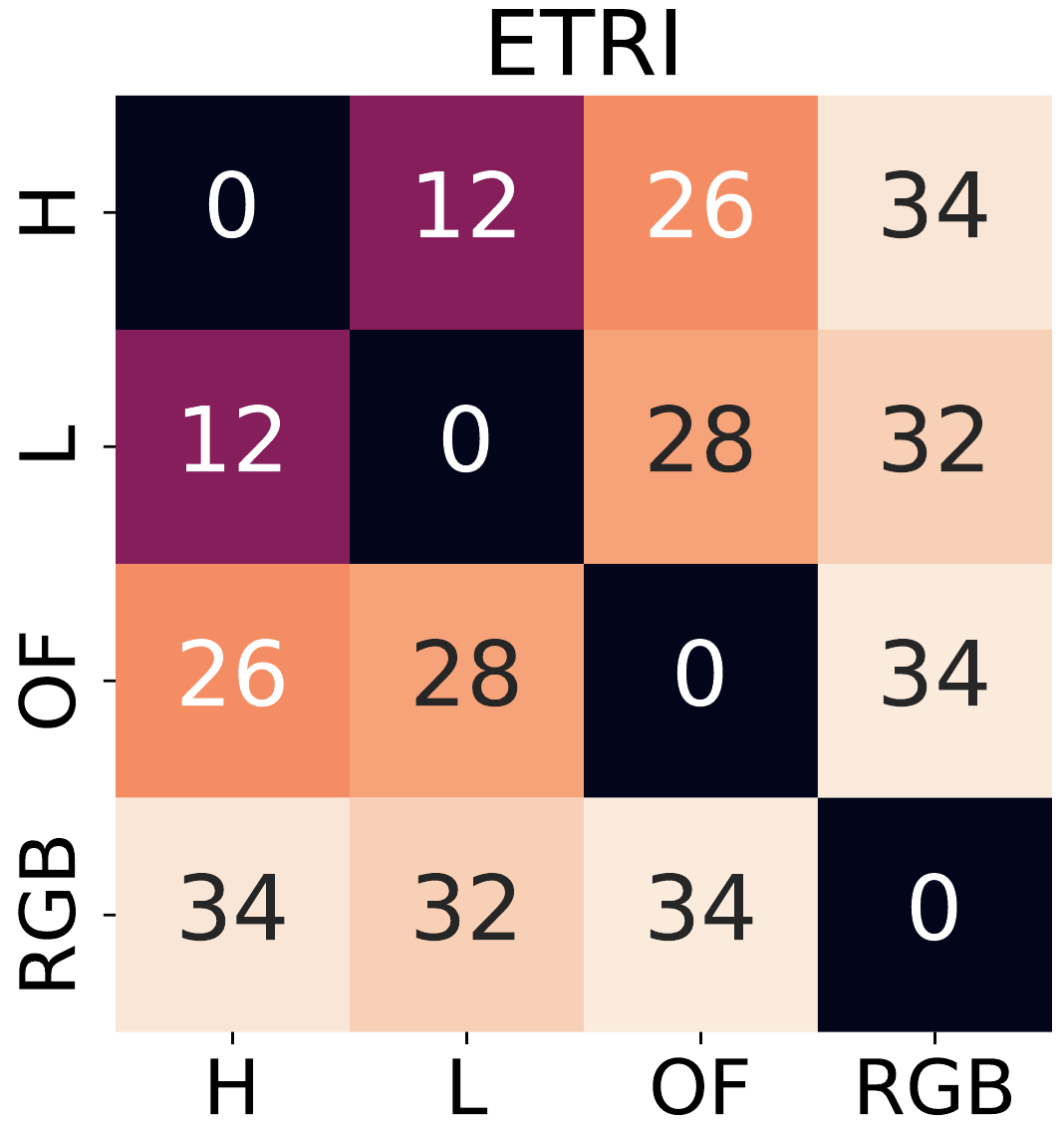}
  }

%\vspace*{-2em}
\caption{Maximum Mean Discrepancy values $MMD(m,n)$ computed with Equation~\ref{eq:mmd_pairs} for all modality pairs  $(m,n)$. Warmer colors correspond to a higher discrepancy.}
\label{fig:mmd_matrices}
\end{figure}

The $MMD(m,n)$ values for all modality pairs in the three test datasets are illustrated in Figure~\ref{fig:mmd_matrices}. The higher discrepancy values are clearly apparent by their bright colors and contrast to the rest of the values. Optical flow has the largest values on Sims4Action, and RGB has the highest discrepancy on Toyota and ETRI. Optical flow also exhibits a high discrepancy on ETRI. Once again, we can see that the domain gap to the ETRI dataset is much larger, which is manifested in drastically higher $MMD(m,n)$ values. The pairs-thresholds $\delta_{\textbf{pair}}^{MMD}$ for the three datasets are $\{13.68, 19.18, 27.50\}$ and separate exactly the same modality pairs as our quantification study and the $\delta_{\textbf{pair}}^{\rho}$ thresholds, with the exception of the (H,OF) pair in ETRI. The aggregated discrepancies $MMD(m)$ for each modality and the aggregated thresholds are listed in Table~\ref{tab:merged_tables}\textbf{(c)}, where the values above the thresholds are colored in red. The red values coincide exactly with the negative contributions $f(m)$ from our quantification study in Table~\ref{tab:merged_tables}\textbf{(a)}.

\textbf{ModSelect: Unsupervised Modality Selection.} Finally, we select the modalities with either the aggregated $\delta_{\textbf{agg}}$ or the pairs-thresholds $\delta_{\textbf{pair}}$, by constructing the consensus between our two metrics $\rho$ and $MMD$ (see Equations~\ref{eq:aggregated_consensus} and~\ref{eq:holistic_consensus}). The selected modalities $\mathcal{M}_{\textbf{agg}}$ with our aggregated thresholds $\delta_{\textbf{agg}}$ and selected modality pairs $\mathcal{C}_{\textbf{pair}}$ with our pairs-thresholds $\delta_{\textbf{pair}}$ are listed in Table~\ref{tab:selected_modalities}. The selected modalities $\mathcal{M}_{\textbf{agg}}$ from our aggregated thresholds $\delta_{\textbf{agg}}$ are exactly the ones with a positive contribution $\mathcal{M}^+$ in Table~\ref{tab:merged_tables}\textbf{(a)} from our quantification study in Section~\ref{sec:baseline_results}. The pairs-thresholds $\delta_{\textbf{pair}}$ have selected only modality pairs $\mathcal{C}_{\textbf{pair}}$ which are constituted out of modalities from $\mathcal{M}^+$, which means that $\mathcal{C}_{\textbf{pair}}$ contains only pairs of modalities $(m,n)$ with positive contributions, i.e. $f(m), f(n) > 0$. In other words, our proposed unsupervised modality selection is able to select only the modalities with positive contributions by utilizing the predictions correlation and MMD between the embeddings of the unimodal action classifiers, without the need of any ground-truth labels on the test datasets.

\textbf{Impact on the multimodal accuracy.} The impact of the unsupervised modality selection on the mean multimodal accuracy can be seen in Table \ref{tab:selected_modalities}. Selecting the modalities with our proposed thresholds leads to an average improvement of $5.2\%$, $3.6\%$, and $4.3\%$ for Sims4Action~\cite{roitberg2021let}, Toyota~\cite{das2019toyota}, and ETRI~\cite{jang2020etri} respectively. This is a substantial improvement, given the low accuracies on the \textsc{Real} test datasets due to the synthetic-to-real domain gap. These results confirm that our modality selection approach is able to discern between good and bad sources of information, even in the case of a large distributional shift.
%The results demonstrate that the average accuracy of our multimodal models can be increased by leveraging the correlation between their predictions as well as the domain discrepancy between their embeddings, both of which require no labels during test time and can be applied in an unsupervised fashion.

\begin{table}[t]
    \centering
    \caption{Results from ModSelect: Selected modalities $\mathcal{M}_{\textbf{agg}}$ with $\delta_{\textbf{agg}}$ and selected modality pairs $\mathcal{C}_{\textbf{pair}}$ with $\delta_{\textbf{pair}}$ computed with Equations~\ref{eq:aggregated_consensus} and~\ref{eq:holistic_consensus}.}
    \label{tab:selected_modalities}
    \scalebox{0.65}{
    \begin{tabular}{l|c|c|c|c}
    \toprule
        {} & {} & {} & \multicolumn{2}{c}{Average multimodal accuracy} \\ \hline
        Test Dataset & $\mathcal{M}_{\textbf{agg}}$ & $\mathcal{C}_{\textbf{pair}}$ & All Modalities $\mathcal{M}$ & Ours: $\mathcal{M}_\textbf{agg}$ \\ \hline 
        Sims4Action~\cite{roitberg2021let} & 
        \{H, L, RGB\} & \multirow{3}{*}{$\{(m,n) \in \mathcal{M}^2 | m \in \mathcal{M}^+ \land n \in \mathcal{M}^+\}$
        } & 85.7\% & 90.9\% \color{darkgreen}{(+5.2\%)} \\ 
        Toyota~\cite{das2019toyota} & \{H, L, OF, YOLO\} & {} & 22.9\% & 26.5\% \color{darkgreen}{(+3.6\%)} \\ 
        ETRI~\cite{jang2020etri} & \{H, L, YOLO\} & {} & 17.7\% & 22.0\% \color{darkgreen}{(+4.3\%)} \\ 
        \bottomrule
    \end{tabular}}
    \vspace*{-0.5cm}
\end{table}

\section{Limitations and Conclusion}
%\iffalse
%Our work introduces a method for quantifying the contribution of each individual modality in a multimodal framework and an additional unsupervised modality selection pipeline.
\textbf{Limitations.} A limitation of our work is that the contributions quantification metric relies on the evaluation results on the test datasets, i.e., the ground-truth labels are needed to calculate the metric. Moreover, the with-without metric $f(m)$ in Equation \ref{eq:contribution} requires $\mathcal{O}(M * 2^{M-1})$ computations, where $M$ is the number of modalities. However, one should note that in practice $M$ is not too large, e.g., $M\leq10$. Additionally, since our method is novel, it has only been tested on the task of cross-domain action recognition. To safely apply our method to other multimodal tasks, e.g., object recognition, future investigations are needed. Moreover, the overall accuracy is still relatively low for cross-domain activity recognition $(<30\%)$ and more research is needed for deployment-ready systems. 
%\fi

\textbf{Conclusion.} This is the first systematic study of modality selection in the context of cross-domain activity recognition, aimed at providing guidance for future work in multimodal domain generalization.
Our experiments validate our assumption, that  cross-domain activity recognition clearly benefits from multimodality, but not all modalities improve the recognition and a systematic modality selection is vital for achieving good results. We proposed a way to measure the contribution of each modality when it is included in a late fusion workflow. The contribution can be used to quantify the importance of each modality and to justify which sources of information are included in a multimodal framework. Our experiments indicate that the correlation between the predictions of unimodal classifiers and the Maximum Mean Discrepancy between their embeddings are both suitable metrics for unsupervised modality selection. The metrics allow to compute thresholds which select only modalities with positive contributions, which opens the possibility to automatically discard bad or uncertain sources of information and to improve the performance on unseen domains. We hope that our findings will provide guidance for a better modality selection process in the future, which is based on more structured and justified decisions.
%Interestingly, while most of the activity recognition research is conducted on RGB videos, this modality consistently under-performed under \textsc{Synthetic$\rightarrow$Real} conditions, often negatively affecting the results if added to a multimodal recognition system.

\mypar{Acknowledgements} This work was supported by the JuBot project sponsored by the Carl Zeiss Stiftung and Competence Center Karlsruhe for AI Systems Engineering (CC-KING) sponsored by the
Ministry of Economic Affairs, Labour and Housing Baden-W{\"u}rttemberg.

\bibliographystyle{splncs04}
\bibliography{bib/egbib,bib/late_fusion_learned,bib/late_fusion_score_rule,bib/late_fusion_driver,bib/domain_generalization,bib/self_supervised_learning,bib/early_fusion,bib/da_maximum_mean_discprepancy}
\end{document}